\documentclass{article}

\PassOptionsToPackage{square,numbers}{natbib}

\usepackage[preprint]{neurips_2021}

\usepackage[utf8]{inputenc} 
\usepackage[T1]{fontenc}    
\usepackage{hyperref}       
\usepackage{url}            
\usepackage{booktabs}       
\usepackage{amsfonts}       
\usepackage{nicefrac}       
\usepackage{microtype}      
\usepackage{xcolor}         

\usepackage[pdftex]{graphicx}
\usepackage{caption}
\usepackage{float}
\usepackage{subcaption}
\usepackage{comment}

\usepackage{listings}
\usepackage{color}
 
\definecolor{codegreen}{rgb}{0,0.6,0}
\definecolor{codegray}{rgb}{0.5,0.5,0.5}
\definecolor{codepurple}{rgb}{0.58,0,0.82}
\definecolor{backcolour}{rgb}{0.95,0.95,0.92}
 
\usepackage[colorinlistoftodos]{todonotes}

\title{Visual Referential Games Further the Emergence of Disentangled Representations 
}

\author{%
  Kevin Denamganaï, Sondess Missaoui, and James Alfred Walker \\
  Department of Computer Science\\
  University of York\\
  York, UK \\
  \texttt{kyd500@york.ac.uk}, \texttt{sondess.missaoui@york.ac.uk}, \texttt{james.walker@york.ac.uk} \\
}  

\begin{document}

\maketitle


\begin{abstract}
    
    Natural languages are powerful tools wielded by human beings to communicate information. Among their desirable properties, compositionality has been the main focus in the context of referential games and variants, as it promises to enable greater systematicity to the agents which would wield it. 
    The concept of disentanglement has been shown to be of paramount importance to learned representations that generalise well in deep learning, and is thought to be a necessary condition to enable systematicity. Thus, this paper investigates how do compositionality at the level of the emerging languages, disentanglement at the level of the learned representations, and systematicity relate to each other in the context of visual referential games.
    Firstly, we find that visual referential games that are based on the Obverter architecture outperforms state-of-the-art unsupervised learning approach in terms of many major disentanglement metrics.
    Secondly, we expand the previously proposed Positional Disentanglement (PosDis) metric for compositionality to (re-)incorporate some concerns pertaining to informativeness and completeness features found in the Mutual Information Gap (MIG) disentanglement metric it stems from. This extension allows for further discrimination between the different kind of compositional languages that emerge in the context of Obverter-based referential games, in a way that neither the referential game accuracy nor previous metrics were able to capture. 
    Finally we investigate whether the resulting (emergent) systematicity, as measured by zero-shot compositional learning tests, correlates with any of the disentanglement and compositionality metrics proposed so far. Throughout the training process, statically significant correlation coefficients can be found both positive and negative depending on the moment of the measure. Thus, our results on that end are inconclusive and it shows that more theoretical work is necessary.  
    
\end{abstract}

\section{Introduction}

Visual referential games are at an interface between the language processing subfields of language emergence, language grounding, and the computer vision subfield of unsupervised representation learning.
While language emergence raises the question of how to make artificial languages emerge with similar properties to natural languages, or at least `natural-like' protolanguages, with compositionality at the forefront of those properties\citep{Baroni2019, Guo2019, Li&Bowling2019, Ren2020}, language grounding is concerned with the ability to ground the meaning of (natural) language utterances into some sensory processes, with the visual modality being the main focus of research. On one hand, emerging artificial languages' compositionality has been shown to further the learnability of said languages \citep{kirby2002learning, Smith2003, Brighton2002, Li&Bowling2019} and, on the other hand, natural languages' compositionality promises to increase the generalisation ability of the artificial agent that would be able to rely on them as a grounding signal, as it has been found to produce learned representations that generalise, when measured in terms of the data-efficiency of subsequent transfer and/or curriculum learning \citep{Higgins2017SCAN, Mordatch2017, MoritzHermann2017, Jiang2019}. More in touch with the current context of this study, \citet{Chaabouni2020} showed that, when a specific kind of compositionality is found in the emerging languages (the kind that scores high on the positional disentanglement (posdis) metric for compositionality that they proposed), then it is a sufficient condition for systematicity to emerge.

Emerging languages are far from being `natural-like' protolanguages \citep{Kottur2017,Chaabouni2019a,Chaabouni2019b}, but sufficient conditions can be found to further the emergence of compositional languages and generalising learned representations (e.g. ~\citet{Kottur2017, Lazaridou2018, Choi2018, Bogin2018, Guo2019, Korbak2019, Chaabouni2020, DenamganaiAndWalker2020b}). Nevertheless, the ability of neural networks to generalise in a systematic fashion has been called into question, especially when it comes to language grounding in general~\cite{Hill2019-tm}, on relational reasoning tasks~\cite{Bahdanau2019}, or on the SCAN benchmark ~\citep{Lake&Baroni2018, Loula2018, Liska2018}, and more recently the gSCAN benchmark~\cite{Ruis2020-vj}. Neural networks induction biases have been investigated towards finding necessary conditions that favour the emergence of systematicity~\citep{Hill2019-tm, Slowik2020, Korrel2019, Lake2019, Russin2019}.  




\textbf{Compositionality \& Systematic Generalisation/Systematicity.} As a concept, compositionality has been the focus of many definition attempts. For instance, it can be defined as ``the algebraic capacity to understand and produce novel combinations from known components''(\citet{Loula2018} referring to \citet{montague1970universal}) or as the property according to which ``the meaning of a complex expression is a function of the meaning of its immediate syntactic parts and the way in which they are combined''~\citep{krifka2001compositionality}. Although difficult to define, the commmunity seem to agree on the fact that it would enable learning agents to exhibit systematic generalisation abilities (also referred to as combinatorial generalisation \citep{Battaglia2018}). 
Some of the ambiguities that come with those loose definitions start to be better understood and explained, as in the work of~\citet{Hupkes2019}. In this paper, we will refer to compositionality as ``the ability to entertain a given thought implies the ability to entertain thoughts with semantically related contents''\citep{Fodor&Pylyshyn1988}, and thus use it interchangeably with systematicity, following the classification made by \citet{Hupkes2019}.

Compositionality, as a property of languages, can be difficult to measure. \citet{Brighton&Kirby2006}'s \textit{topographic similarity} (\textbf{topsim}) which is acknowledged by the research community as the main quantitative metric for compositionality~\citep{Lazaridou2018, Guo2019,Slowik2020, Chaabouni2020, Ren2020}. Recently, taking inspiration from disentanglement metrics, \citet{Chaabouni2020} proposed two new metrics entitled \textbf{posdis} (positional disentanglement metric) and \textbf{bosdis} (bag-of-symbols disentanglement metric), that have been shown to be differently `opinionated' in the sense that they each seem to capture different ways in which a language can be shown to be compositional. 

In accordance with \citet{Hill2019-tm}, \citet{Chaabouni2020} also found that compositionality is not necessary but only sufficient to bring about systematic generalisation, as shown by the fact that non-compositional languages wielded by symbolic (generative) referential game players were enough to support systematic generalisation as evaluated by \textit{zero-shot compositional learning tests} (in which a set of stimuli composed of specific attributes are held-out from the training set, while making sure that the agents are still familiarising themselves with the specific attributes in different contexts/combinations to the held-out ones). One of the necessary conditions they found to foster generalisation is the richness of stimuli (i.e. many possible values per attribute/latent dimension). Therefore, in this paper, on top of measuring the different ways in which emerging languages can be compositional, using topsim, posdis, and bosdis, we will perform zero-shot compositional learning tests in the context of a \textbf{``poorness'' of stimuli}, in order to measure systematic generalisation abilities/systematicity in the most difficult context. 

\textbf{Disentanglement \& Compositionality in Visual Referential Games.} Disentanglement, as a property of learned representations, is the objective of the unsupervised representation learning field. It has been shown to be worthy of pursuit for disentangled learned representations are, at least, more sample-efficient and, at best, enabling greater performance, when considering subsequent upstream tasks~\cite{Higgins2017SCAN, Higgins2017DARLA, Van_Steenkiste2019-xm}. Nevertheless, it also suffers from a definition issue~\cite{Higgins2018,Eastwood_2018-cr} that spawned many different metrics with different sensibility and correlations among themselves~\cite{Locatello2020-cx}. 

In the context of visual referential games (as opposed to symbolic ones), we are entitled to wonder \textbf{(i) in the objective of promoting compositionality in the emerging languages, how do disentangled learned representations relate to the emergence of compositionality}, \textbf{(ii) from the perspective of unsupervised representation learning, whether some induction biases of referential games promote disentanglement}, and \textbf{(iii) how do disentanglement and compositionality relate with (emergent) systematicity}.

Therefore, in order to measure disentanglement, we will make use of the \textbf{FactorVAE Score}~\cite{Kim2018}, the Mutual Information Gap (\textbf{MIG})~\cite{Chen_2018-MIG}, and the \textbf{Modularity Score}~\cite{Ridgeway2018-xo} as they have been shown to be part of the metrics that correlate the least among each other by \citet{Locatello2020-cx}. Moreover, the \textbf{posdis} and \textbf{bosdis} metrics for compositionality have been inspired by the \textbf{MIG} metric for disentanglement, therefore it is relevant to verify whether compositionality as measured by the former is correlated with disentanglement as measured by the latter. By including many metrics for both disentanglement and compositionality, we aim to make sure to capture any correlation between the many opinionated definitions of the concepts of disentanglement and compositionality.

\textbf{Contributions.} Firstly, following \citet{Chaabouni2020}'s introduction of the posdis and bosdis metrics inspired by the MIG disentangelement metric, we expand on their work by developing a different version, more opinionated in the sense that the compositionality that it highlights is imbued with some disentanglement concerns, we discuss the advantages and disadvantages of both versions via the lenses of decoding and compression in Section~\ref{sec:metrics}. Our version finds its motivation in the need to discriminate further between the different kind of compositional languages that emerges in the context of obverter-based referential games, as detailed in Section~\ref{subsec:kind-of-compo}.

Secondly, we show experimental results in Section~\ref{subsec:rg-for-disent} that the visual referential game paradigm furthers strong disentanglement at the level of learned representations, outperforming the state-of-the-art method FactorVAE. This paradigm is thus highlighted as an interestingly viable alternative to VAE-based~\cite{Higgins2016,Kim2018,Chen_2018-MIG} and GAN-based~\cite{chen2016infogan} approaches to the unsupervised learning of disentangled representations, in the context of a ``poorness'' of the stimuli. 

Finally we investigated whether the resulting (emergent) systematicity, as measured by zero-shot compositional learning tests, correlates with any of the disentanglement and compositionality metrics proposed so far, in the context of visual referential games (see Section~\ref{subsec:dis-compo-system}). Throughout the training process, statistically significant correlation coefficients can be found as both positive or negative depending on the timing of the measure. Thus, our results on that end are inconclusive, and it shows that more theoretical work is necessary.

\section{Setup}
\subsection{Visual Referential Games}
\label{sec:visual-referential-games}
\begin{figure*}[t]
    \centering
    \begin{subfigure}[t]{0.62\linewidth}
        \includegraphics[width=1.0\linewidth]{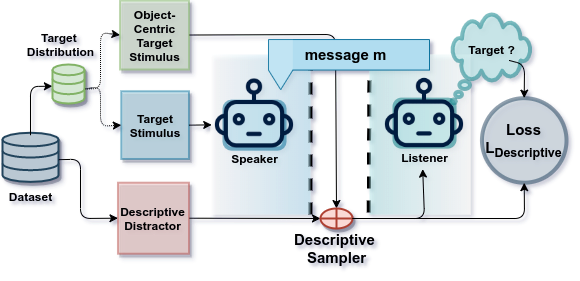}
        \caption{}
        \label{fig:referentialgame}
    \end{subfigure}
    \begin{subfigure}[t]{0.37\linewidth}
        \includegraphics[width=1.0\linewidth]{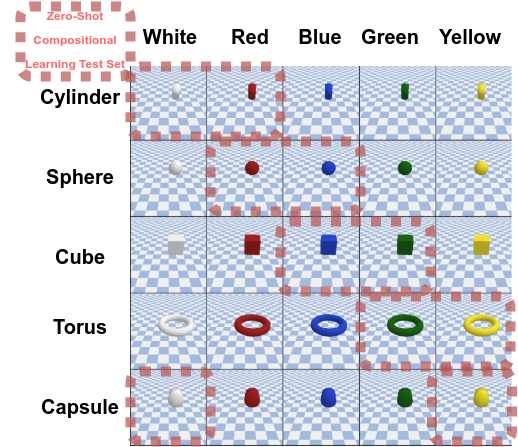}
        \caption{}
        \label{fig:dataset-example}
    \end{subfigure}
    \caption{\textbf{(a):} Illustration of a \textit{descriptive object-centric (partially-observable) $2$-players/$L$-signal/$N=0$-round/$K=0$-distractor} \textit{referential game}. 
    \textbf{(b):} Example images of the dataset showing the different objects (in specific viewpoints), along with a possible zero-shot compositional learning train-test split example.
    }
\end{figure*}
Referential/Language games emphasise the functionality of languages, namely, the ability to efficiently communicate and coordinate between agents. Following the nomenclature proposed in \citet{Denamganai2020a}, we will focus primarily on a \textit{descriptive object-centric (partially-observable) $2$-players/$L=10$-signal/$N=0$-round/$K=0$-distractor} referential game variant, as illustrated in Figure~\ref{fig:referentialgame}, with a descriptive ratio of $0.5$. This ratio stands for the probability of the target stimuli to be shown to the \textit{listener} agent, while the rest of the time it is substituted for the \textit{Descriptive Distractor}(see Figure~\ref{fig:referentialgame}).

As an object-centric referential game, as opposed to stimulus-centric, the listener and speaker agents are not being presented with the very same target stimuli. Rather, they are being presented with different \textit{viewpoints} on the very same target object shown in the target stimuli, where the word \textit{viewpoint} ought to be understood in a very large sense. Indeed, object-centrism is implemented by designing one of the latent axes of the dataset as a source of invariance for the concept of object, the \textbf{object-centric latent axis}. Thus, the listener and speaker agents would be presented with different stimuli that keeps the conceptual object being presented constant, i.e. that keeps constant the values on all the other latent axes but the object-centric latent axis value.
This aspect was introduced by \citet{Choi2018} (without it being of primary interest), where the pair of agents would literally be shown potentially the same 3D objects under different viewpoint, thus thinking of object-centrism as a \textit{viewpoint} shift is historically relevant.

Concerning the communication channel, the vocabulary $V$ is fixed with $10$ ungrounded symbols, plus an eleventh grounded symbol accounting for the \textit{end of sentence} semantic, thus $|V|=11$. The maximum sentence length $L$ is always equal to $10$, thus placing our experiments in the context of an overcomplete communication channel whose capacity is far greater than the number of different meanings that the agents would encounter in our experiments.

We will focus exclusively on two parameterisation of the communication channel. Fisrtly, on the \textit{Straight-Through Gumbel-Softmax} (STGS) approach proposed by \citet{Havrylov2017}, as it supposedly allows a richer signal towards solving the credit assignment problem that language emergence poses since the gradient can be backpropagated from the listener agent to the speaker agent, while, in comparison, it cannot be backpropagated when using more commonly adopted approaches based on REINFORCE-like algorithms~\citep{williams1992simple}. 
And, secondly, we investigate the \textit{Obverter} approach proposed by~\citet{batali1998computational} and updated to the recent deep learning paradigm by~\citet{Choi2018, Bogin2018}, for its induction bias has proven itself powerful. As a well-discussed concept in the Theory of Mind, it posits that the speaker agent should make the assumption that the listener's mind operates similarly to its own, thus using its own understanding of a given utterance as a good enough proxy of the expected listener's understanding of the same utterance, when trying to convey a given meaning in the obvious constraint that the listener's state of mind is inaccessible to it.

Having described the referential game setup, the following section provides details on the architecture of the \textit{speaker} and \textit{listener} agents and the dataset used\footnote{For more details, please refer to our code released at: \url{https://github.com/Near32/ReferentialGym/tree/develop/zoo/referential-games\%2Bcompositionality\%2Bdisentanglement}.}.

\subsection{Agent Architectures}
\label{subsec:arch}

Each agent consist of, at least, a language module and a visual module. The \textit{listener} agents also incorporates a third decision module that combines the outputs of the other two modules. In the case of the Obverter approach, both agents play the role of the \textit{listener} from one round to another, therefore they both incorporate it. \textcolor{red}{As this work focuses on learning `good' representations, we make the architectural choice of sharing the visual module between the pairs of agents, as preliminary experiments showed it increases sample-efficiency.}

In the case of the STGS approach, while the \textit{speaker} agent is prompted to produce the output string of symbols with a \textit{Start-of-Sentence} symbol and the visual module output as an initial hidden state, the \textit{listener} agent consumes the string of symbols with the null vector as the initial hidden state. In the following subsections, we detail each module architecture in depth.

\textbf{Visual Module.} The visual module $f(\cdot)$ consists of four $3\times3$ convolutional layers with stride $2$, followed by a fully-connected layer reducing the feature maps to flattened vectors of dimension $32$. The two first convolutional layers have $32$ filters, whilst the last two layers have $64$. Each convolutional layer is followed by a $2$D batch normalisation layer, which are found crucial in the case of the Obverter approach (without it, training does not take off), and the resulting outputs are passed through ReLU activation functions. The bias parameters of the convolutional layers are not used, as it is common when using batch normalisation layers. Inputs are resized to $64\times64$, thus yielding feature maps of dimension $64\times4\times4$. The input to the final fully-connected layer is a flattened representation of dimension $1024$. 

\textbf{Language Module.} The language module $g(\cdot)$ consists of a one-layer GRU network~\citep{cho2014learning} in the case of the Obverter approach, and a one-layer LSTM network~\citep{hochreiter1997long} in the case of the STGS, with $64$ hidden units. 
In the context of the \textit{listener} agent, the input message $m=(m_i)_{i\in[1,L]}$ (produced by the \textit{speaker} agent) is represented as a string of one-hot encoded vectors of dimension $|V|$ and embedded in an embedding space of dimension $64$ via a learned Embedding. The output of the \textit{listener} agent's language module, $g^l(\cdot)$, is the last hidden state of the LSTM layer or GRU layer, $h^l_L = g^l(m_L, h^l_{L-1})$.
In the context of the \textit{speaker} agent's language module $g^s(\cdot)$, the output is the message $m=(m_i)_{i\in[1,L]}$ consisting of one-hot encoded vectors of dimension $|V|$, which are sampled using the STGS approach from a categorical distribution $Cat(p_i)$ where $p_i = Softmax(\nu(h^s_i))$, provided $\nu$ is an affine transformation and $h^s_i=g^s(m_{i-1}, h^s_{i-1})$. $h^s_0=f(s_t)$ is the output of the visual module, given the target stimulus $s_t$.

\textbf{Decision Module.} Depending on the approach, the decision module can be very different. In the case of the STGS, similarly to \citet{Havrylov2017}, the decision module builds a probability distribution over a set of $K+1$ stimuli/images $(s_0, ..., s_K)$, consisting of $K$ distractor stimuli and the target stimulus (or possibly another distractor in the case of the descriptive games), given a message $m$:
\begin{equation}
\label{eq:discr-stgs}
    p((d_i)_{i\in[0,K]}| (s_i)_{i\in[0,K]} ; m) = Softmax( ( h^l_L \cdot f(s_i)^T )_{i\in[0,K]} ).
\end{equation}

In our case though, since there are no distractors, we set $K=0$, and, since the agents play a descriptive game, \textcolor{blue}{a final category is added to encode the meaning/prediction that none of the $K+1$ stimuli is the target stimulus that the \textit{speaker} agent was `talking' about}. The addition is made at the logit level as a \textcolor{red}{learnable logit value, $logit_{no-target}$}, it is an extra parameter of the model. In this case the decision module output is as follows:
\begin{equation}
\label{eq:descr-discr-stgs}
    p((d_i)_{i\in[0,\textcolor{blue}{K+1}]} | (s_i)_{i\in[0,K]}; m) = Softmax( ( h^l_L \cdot f(s_i)^T )_{i\in[0,K]} \cup \{\textcolor{red}{logit_{no-target}}\} ),
\end{equation}
where $p(d_{\textcolor{blue}{K+1}} | (s_i)_{i\in[0,K]};m)$ is the predicted likelihood that none of the experienced stimuli is the target stimulus. \\

Similarly to \citet{Choi2018}, the decision module of Obverter agents is a two-layer fully-connected network $d(\cdot)$ with $128$ hidden units and a ReLU activation function, taking as input the concatenation of the outputs of the visual and language modules, and outputting $2$ logits for each input stimulus. The first logit $d^{0}(s_i)$ encodes that the listener agent is experiencing the same stimulus than the speaker agent, while the second one $d^{1}(s_i)$ encodes the negation. Given a set of $K+1$ stimuli, the prediction distribution of the listener agent is as follows:

\begin{equation}
\label{eq:descr-discr-obverter}
    p((d_i)_{i\in[0,\textcolor{blue}{K+1}]} | (s_i)_{i\in[0,K]} ; m) = Softmax( ( d^{0}(s_i) )_{i\in[0,K]} \cup \{\textcolor{red}{\frac{1}{K+1}\sum_{i=0}^{K}{d^{1}(s_i)}}\}).
\end{equation}
We also experimented with taking the minimum and maximum of the set of logits that encodes the negation, but the most efficient approach is by using an \textcolor{red}{average pooling over the set of negation-encoding logits}, as presented in Equation~\ref{eq:descr-discr-obverter}. \\

\subsection{Dataset}
\label{subsec:datasets}

In the following experiments, learning agents observe visual stimuli from particular train/test splits of a replication of the dataset used in the original work of \citet{Choi2018}, that we will refer to as 3DShapesPyBullet, and that we open-source (HIDDEN-FOR-REVIEW-PURPOSE).



Built in order to be employed as a benchmark for disentangled representation learning, the 3DShapesPyBullet dataset consists of visual representations of objects with the following attributes (illustrated in Figure~\ref{fig:dataset-example}): a \textbf{Shape} attribute with $5$ different values; a \textbf{Color} attribute with $5$ different values; a \textbf{Viewpoint} attribute with $10$ different values.
All combinations of values along any of the generative factors/attributes/latent axes are part of the dataset, thus yielding a dataset of size $250$, which fit in the context of the \textbf{``poorness'' of the stimuli}. 
Using an object-centric referential game, we define the \textbf{Viewpoint} latent axis as our \textbf{object-centric latent axis}.





In order to build a zero-shot compositional learning test set, half of the possible values for each attribute are defined as testing-purpose values. 
Subsequently, for every value on the \textbf{Shape} latent axis, $2$ out of the $5$ available \textbf{Color} values are defined as testing-purpose values for this specific \textbf{Shape} value and the resulting stimuli are held-out, as shown in Figure~\ref{fig:dataset-example}.
Rather than sampling the held-out \textbf{Color} values to a given \textbf{Shape} value in a random fashion, we make sure that they are different from one \textbf{Shape} value to another and that all the possible \textbf{Color} values are represented at least with one \textbf{Shape} value in the training set.
This ensures that the agent is familiar with all the possible values on each latent axis while remaining unaware of a subset of all the combinations possible.

The choice of defining around half of the possible values on each attribute is motivated by the results of \citet{Bahdanau2019} when evaluating for emergent systematicity: a CNN+LSTM architecture, similar to that of ours here, has been found sufficiently challenged by such a train/test split.



In the original work of \citet{Choi2018}, a specific batch sampling scheme was used in order to guide the language emergence process towards greater completeness and informativeness (see Section~\ref{subsec:mig-posdis}), by sampling batches of stimuli with controlled amount of \textit{Descriptive Distractors} that would differ from the target exclusively in terms of the \textbf{Shape} or the \textbf{Color} attribute value, in a form of supervised contrastive learning. \textcolor{red}{We do not make use of this sampling scheme as we place ourselves in the context of unsupervised learning, where the actual latent factors should not be known in advance.}



\section{Metrics}
\label{sec:metrics}

In order to measure disentanglement, we adapted to our framework the implementations of the \textbf{FactorVAE Score}~\cite{Kim2018}, the Mutual Information Gap (\textbf{MIG})~\cite{Chen_2018-MIG}, and the \textbf{Modularity Score}~\cite{Ridgeway2018-xo} from the open-source work of \citet{Locatello2020-cx}. The choice of those metrics is motivated by their results showing that they correlate the least among each other, which we therefore understand as being differently opinionated about what it means for learned representations to be called disentangled.

By incorporating those different opinions/definitions about both disentanglement and compositionality, we aim to answer more thoroughly the question of finding whether disentanglement, compositionality, and systematicity are linked in the context of (discriminative) referential games. 
Moreover, the \textbf{posdis} and \textbf{bosdis} metrics for compositionality have been inspired by the \textbf{MIG} metric for disentanglement, therefore it is relevant to verify whether compositionality as measured by the former is correlated with disentanglement as measured by the latter. By including many metrics for both disentanglement and compositionality, we aim to make sure to capture any correlation between the many opinionated definitions of the concepts of disentanglement and compositionality.

\subsection{Positional Disentanglement Metrics Formulation}
\label{subsec:mig-posdis}

As detailed in \citet{Chen_2018-MIG}, the Mutual Information Gap (MIG) metric is defined for \textcolor{blue}{latent variables $(z_j)_{j\in [1,J]}$} and \textcolor{red}{ground truth factors $(v_k)_{k\in [1;K]}$}, where $J$ is the dimension of the model at the level of its latent variables and $K$ is the number of factors in the current dataset. The MIG ``enforce[s] axis-alignment by measuring the difference between the top two latent variables with highest mutual information'', for each ground truth factor. Summing over all ground truth factors yield the MIG score:
\begin{equation}
    MIG(\textcolor{blue}{z},\textcolor{red}{v}) = \frac{1}{K}\sum_{k=1}^K \frac{1}{\mathcal{H}(\textcolor{red}{v_k})} \Big{(} \mathcal{I}(\textcolor{blue}{z_{j_k^{(1)}}}; \textcolor{red}{v_k}) - \mathcal{I}(\textcolor{blue}{z_{j_k^{(2)}}}; \textcolor{red}{v_k})  \Big{)}
    \label{eq:mig}
\end{equation}
where $j_k^{(l)}$ is the index of the latent variable with the $l$-th highest mutual information with the $k$-th ground truth factor, for instance $\textcolor{blue}{z_{j_k^{(1)}}} = argmax_{z_j} \mathcal{I}(\textcolor{blue}{z_j} ; \textcolor{red}{v_k)}$, and $\mathcal{H}(\textcolor{red}{v_k})$ refers to the entropy of the $k$-th ground truth factor and is used for normalization purpose since $\forall (j,k)\in [1,J]\times[1,K],\, 0 \leq \mathcal{I}(\textcolor{blue}{z_j}, \textcolor{red}{v_k}) \leq \mathcal{H}(\textcolor{red}{v_k})$.

On the other hand, the Positional Disentanglement (posdis) metric is defined for \textcolor{blue}{symbol positions within sentences $(s_j)_{j\in [1,c_{len}]}$} and, similarly to the MIG case, \textcolor{red}{ground truth factors $(v_k)_{k\in [1;K]}$}, where $c_{len}$ is the length of sentences uttered by the speaker of consideration (when omitting symbol positions with zero entropy). The posdis is defined, on the contrary to MIG, for each \textcolor{blue}{symbol position $s_j$ within sentences}, and summed over those possible symbol positions that are not constant (non-null entropy), rather than over the \textcolor{red}{ground truth factors}:
\begin{equation}
    posdis(\textcolor{blue}{s},\textcolor{red}{v}) = \frac{1}{c_{len}}\sum_{j=1}^{c_{len}} \frac{1}{\mathcal{H}(\textcolor{blue}{s_j})} \Big{(} \mathcal{I}(\textcolor{red}{v_{k_j^{(1)}}}; \textcolor{blue}{s_j}) - \mathcal{I}(\textcolor{red}{v_{k_j^{(2)}}}; \textcolor{blue}{s_j})  \Big{)}
    \label{eq:posdis}
\end{equation}
where $k_j^{(l)}$ is the index of the ground truth factor with the $l$-th highest mutual information with the $j$-th symbol position within sentences, for instance $\textcolor{red}{v_{k_j^{(1)}}} = argmax_{v_k} \mathcal{I}(\textcolor{red}{v_k} ; \textcolor{blue}{s_j)}$. 

Given that posdis and bosdis are defined in the context of a generative referential game where the listener agent aims to reconstruct the ground truth factors, it can be argued that these metrics take the viewpoint of the listener agent who attempts to decode the speaker's sentences in order to output predicted factors (aiming to converge on the ground truth factors). 
In other words, the philosophy of theses metrics is that of a decoding problem rather than that of a compression problem.

The compression approach would revolve around the speaker agent's viewpoint as it aims to compress the observed stimuli, that reflects the ground truth factors (or are the ground truth factors themselves in the case of symbolic referential games), into \textbf{informative}, \textbf{complete} and hopefully compositional sentence utterances. To further flesh out the parallel with some disentanglement definitions~\cite{Eastwood_2018-cr, Ridgeway2018-xo}, one could easily swap compositional in the last sentence for \textbf{modular}. Thus, a speaker-centred positional disentanglement metric would straightforwardly apply the formulae of Equation~\ref{eq:mig}:
\begin{equation}
    speaker-posdis(\textcolor{blue}{s},\textcolor{red}{v}) = MIG(\textcolor{blue}{s},\textcolor{red}{v}) = \frac{1}{K}\sum_{k=1}^K \frac{1}{\mathcal{H}(\textcolor{red}{v_k})} \Big{(} \mathcal{I}(\textcolor{blue}{s_{j_k^{(1)}}}; \textcolor{red}{v_k}) - \mathcal{I}(\textcolor{blue}{s_{j_k^{(2)}}}; \textcolor{red}{v_k})  \Big{)}
    \label{eq:speaker-posdis}
\end{equation}
The main advantage of the speaker-centred formulation over the listener-centred one is twofold. Firstly, it is able to discriminate between complete and incomplete languages, following the definition of completeness by \citet{Eastwood_2018-cr}: ``the degree to which each underlying factor is captured by a single code variable'', or, in our case, a single symbol position within sentences. For instance, a language that would capture some underlying factor $v_{k_1}$ using two or more symbol positions within sentences, for instance in a redundant fashion, would yield a lower speaker-centred posdis than languages that would capture it with only one. In the context of cheap talk where there is no cost to communication symbol usage, it might be of lesser importance, but using both metrics in concert would at least enable more insights into the kind of language that emerged.

Secondly, most importantly, the speaker-centred formulation cares about the informativeness (``the amount of information that a representation captures about the underlying factors of variation''\cite{Eastwood_2018-cr}) of the language with respect to \textbf{all} the ground truth factors. Indeed, while the mutual information $\mathcal{I}(\,;\,)$ is symmetric, the posdis and MIG scores are not, and especially when it comes to informativeness: for instance, a language whose sentences would consist of only one symbol position with non-null entropy used to capture \textbf{only one of the many} ground truth factors would score perfectly on the listener-centred posdis formulation, but poorly on the speaker-centred formulation because the other underlying ground truth factor have not been captured.

In terms of disadvantage, the speaker-centred formulation of posdis loses the relaxation made by the listener-centred posdis when compared to topographic similarity: the speaker-centred posdis would penalise languages that do not exhibit a one-to-one attribute-position mapping. Borrowing from \citet{Chaabouni2020}, the different metrics, topsim, bosdis, listener-centred posdis, and speaker-centred posdis are differently opinionated about what it means for a language to be compositional, and we argue that the speaker-centred formulation is one of the most opinionated so far. Using them in concert is thus allowing more insights about the kind of compositionality that may emerge in each of the artificial languages we may consider.

\begin{figure*}[t]
    \centering
    \begin{subfigure}[c]{0.25\linewidth}
        \includegraphics[width=1.0\linewidth]{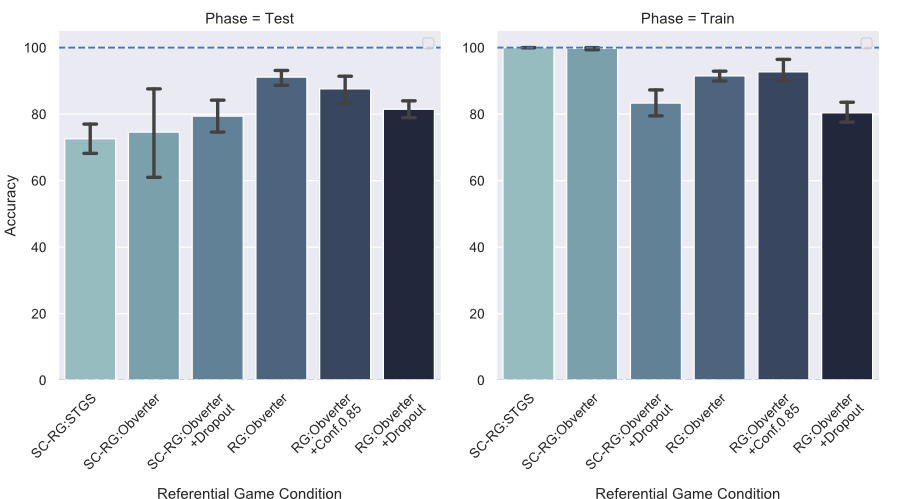}
        \caption{}
        \label{fig:barplot-train-vs-test-3dshapes-low}
    \end{subfigure}
    \begin{subfigure}[c]{0.17\linewidth}
        \includegraphics[width=1.0\linewidth]{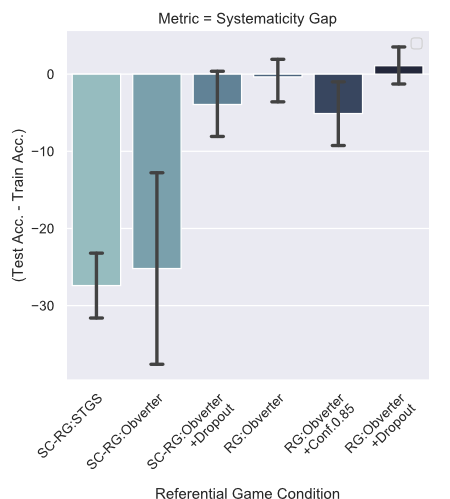}
        \caption{}
        \label{fig:systematicity-gap-3dshapes-low}
    \end{subfigure}
    \begin{subfigure}[c]{0.56\linewidth}
        \includegraphics[width=1.0\linewidth]{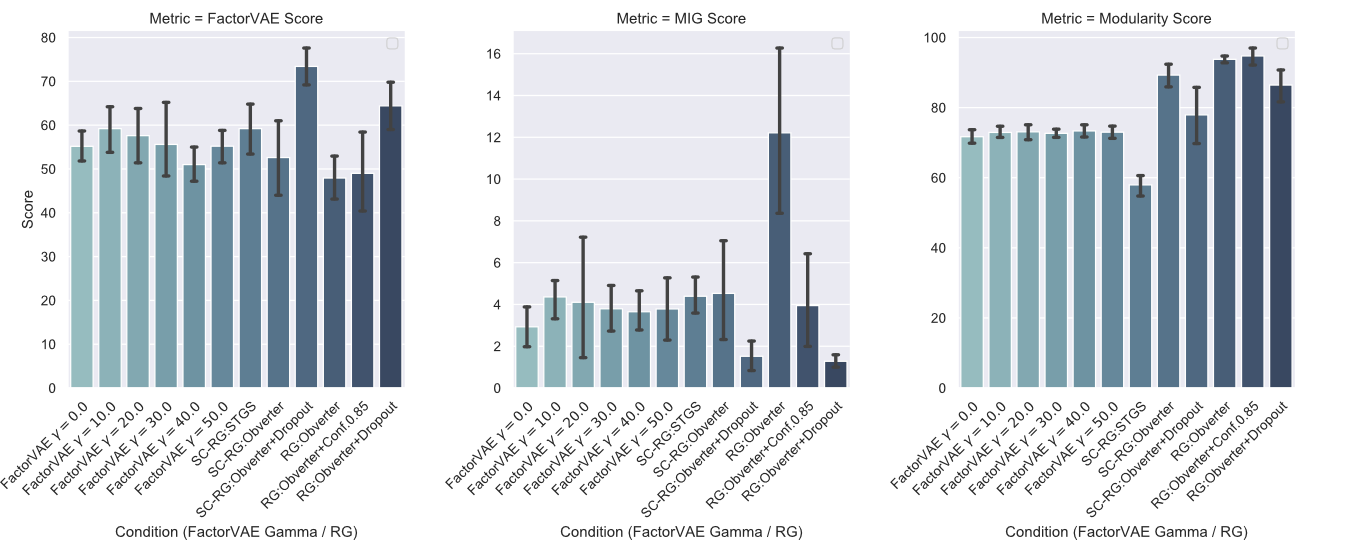}
        \caption{}
        \label{fig:disentanglement-scores-3dshapes-low}
    \end{subfigure}
    \caption{
    Zero-shot compositional learning testing and training accuracy \textbf{(a)} and Systematicity gap (\textbf{(b)}:difference between accuracy on test and train sets) for the different referential game approaches, at the end of training. 
    \textbf{(c)}: Disentanglement scores of the different approaches. The FactorVAE approach is outperformed by the different Obverter-based referential game approaches. 
    }
\end{figure*}

\section{Results}
\label{sec:results}
In this section, we detail our results when training pairs of agents for $4,000$ epochs on the 3DShapesPyBullet dataset, with $5$ random seeds for the contexts with the synthetically compositional communication channels (SC-RG) and between $5$ and $15$ random seeds for the normal referential game contexts (RG). We used the Adam optimizer~\cite{kingma2014adam} with a learning rate of $6.0e^{-4}$ and PyTorch's~\cite{pytorch-paszke-NEURIPS2019_9015} default hyperparameters. The batch size was $64$, as proposed in \citet{Kim2018}. Thus, in the case of the Obverter-based pairs of agents, the role alternation period was $2$.

\subsection{Obverter vs. STGS}
\label{subsec:obv-vs-stgs-dis-and-compo}
Figure~\ref{fig:barplot-train-vs-test-3dshapes-low} and~\ref{fig:systematicity-gap-3dshapes-low} detail at the end of training the referential game accuracy against the training set and the zero-shot compositional learning test set, and the systematicity gap, respectively, for both Obverter-based and STGS-based pairs of agents. Focusing on the SC-RG contexts, we can argue that even when the language is compositional in the sense of the posdis metric, the Obverter-based approaches outperform the STGS-based approach in terms of systematicity (higher absolute accuracies shown in Figure~\ref{fig:barplot-train-vs-test-3dshapes-low}, and less negative gap shown in Figure~\ref{fig:systematicity-gap-3dshapes-low}). Going forward, we focus on the Obverter-based approaches.

\subsection{Referential Games further Disentanglement}
\label{subsec:rg-for-disent}
Figure~\ref{fig:disentanglement-scores-3dshapes-low} details the disentanglement scores of the different approaches. Even when fine-tuning the FactorVAE's hyperparameter $\gamma$, the different Obverter-based referential game approaches outperform the state-of-the-art FactorVAE approach.

We observe a trade-off between the MIG and FactorVAE scores depending on whether a Dropout layer is added just before the Decision modules of the Obverter-based approaches. It highlights the Dropout mechanism as a driver for FactorVAE-centred disentanglement.

Comparing the results when using a synthetically compositional communication channel (SC-RG) from the results when using the normal referential game setting (RG), it is striking to see that disentanglement in the sense of the MIG score is maximized in the RG context, while all the other disentanglement metrics are maximised in the SC-RG contexts, when comparing to respective architecture/approach in the RG context. This result seems to imply that, in the absence of Dropout layers, there is some specific regularisation effect taking place in the Obverter-based RG setting that promotes the emergence of disentanglement in the sense of the MIG score.

\subsection{Compositional Language does not further Listeners' Systematicity}
\label{subsec:compo-not-4-system}
In parallel to the results found in \citet{Chaabouni2020} (showing that when the emerging language is compositional enough in the sense of the posdis metric then the agents wielding it are highly likely to have systematic abilities),  Figure~\ref{fig:systematicity-gap-3dshapes-low} shows that, in spite of learning from a posdis-biased compositional language, neither the listener from an Obverter-based approach or from a STGS-based approach are able to bridge the systematicity gap. 

\textbf{Of the Properties of Symbolic Stimuli.} The main difference between our context and theirs is that they were using symbolic stimuli, while our results is in the context of visual stimuli. Therefore, it could be argued that a specific property found in symbolic stimuli might be a necessary condition for systematicity to emerge when the emerging languages are compositional in a posdis fashion. 

Similarly to \citet{montero2021the}'s interpretation of \citet{Chaabouni2020}'s work, we were expecting this specific property to be captured within one of the disentanglement metrics. Yet, our disentanglement scores found in the context of synthetically compositional communication channels (see Fig.~\ref{fig:disentanglement-scores-3dshapes-low}) are differently high depending on the approach. Of particular interest, the presence of the Dropout layer just before the Decision modules of the Obverter-based agents was highlighted earlier as a mechanism furthering high disentanglement of the learned representations in the sense of the FactorVAE score, and, in the case of the systematicity gap, it induces the least negative systematicity gap among the SC-RG contexts, and the most positive systematicity gap among the RG settings. 

Thus, our results would push forward the idea that, on top of promoting generalisation in the broader sense, the addition of specific Dropout layers in the architecture also brings about systematicity (/compositional or algebraic generalisation), in the context of visual referential games, and high disentanglement of learned representations in the sense of the FactorVAE score is correlated to high systematicity.

\textbf{Of the Importance of the Language Emergence Process.} It is important to also highlight the limitation that our SC-RG contexts do not allow for the non-stationary process through which the language used by the pair of agents would emerge as highly compositional in the sense of the posdis metric, on the contrary to \citet{Chaabouni2020}'s protocol. Therefore, both the stimuli nature (symbolic or visual), on one hand, and the presence or lack there off of the language emergence process, on the other hand, could be necessary conditions for the emergence of systematicity, when the language is already compositional in the sense of the posdis metric. 


\begin{figure*}[t]
    \centering
    \begin{subfigure}[t]{1.0\linewidth}
        \includegraphics[width=1.0\linewidth]{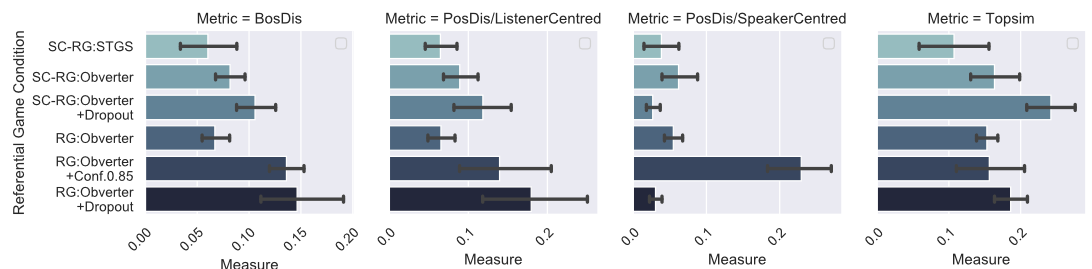}
    \end{subfigure}
    \caption{
    Compositionality metrics for the different referential game approaches, at the end of training.  
    }
    \label{fig:compositionality-scores-3dshapes-low}
\end{figure*}

\subsection{Disentangling Compositionality in Emerging Languages?}
\label{subsec:kind-of-compo}

Figure~\ref{fig:compositionality-scores-3dshapes-low} details the different attributes of the emerging languages in the different referential game approaches. The comparison of the speaker-centred and listener-centred posdis results in the RG contexts highlights that reducing the Obverter's confidence threshold (from the default value of $0.98$ to $0.85$) significantly affects the informativeness and completeness of the emerging language, enabling us to discriminate between further kind of positionally disentangled compositionality.

Recalling the MIG scores in the RG contexts from Figure~\ref{fig:disentanglement-scores-3dshapes-low}, we can see that high (or low) disentanglement in the sense of the MIG score metric does not correlate with high (or low) compositionality in the sense of any posdis metrics. This result subsequently thickens the mystery behind disentanglement and compositionality's relationship.

\subsection{Disentanglement, Compositionality, \& Systematicity are not Naïvely Correlated}
\label{subsec:dis-compo-system}


We compute Spearman $\rho$ correlation matrices between the different metrics throughout the training process of 
Obverter-based approaches (see Appendix~\ref{sec:appendixA}), from epoch $200$ to $4000$, with an interval of $200$ epoch between measures. While some statistically significant coefficients can be found throughout the training process, they are not consistent and rather often contradictory, irrespective of the context. 
From those results, the best we can conclude is that our current concepts for disentanglement, compositionality, and systematicity do not naïvely correlate. We leave it to future works to investigate analysis of a different kind, for instance as time-series variables.

\section{Conclusion}

Primarily, we found that Obverter-based visual referential games outperform the state-of-the-art unsupervised learning FactorVAE approach in terms of many major disentanglement metrics, and highlighted the necessary design choices that lead to that result.
Secondly, we expanded the posdis metric in a principled way with respect to disentanglement concepts of informativeness and completeness, thus allowing for further discrimination between the different kind of compositional languages that emerge in the context of referential games. 
Finally, investigating the relationship between disentanglement, compositionality and systematicity in this unique paradigm of visual referential games, our results provide further evidence that compositionality, disentanglement, and systematicity are not related as expected and are, thus, not fully understood yet.

\section*{Broader Impact}

This work consists solely of simulations, thus alleviating some of the ethical concerns, as well as concerns regarding any consequences emerging due to the failure of the system presented. With regards to the ethical aspects related to its inclusion in the field of Artificial Intelligence, we argue that our work aims to have positive outcomes on the development of human-machine interfaces, albeit being not yet mature enough to aim for this goal. The current state of our work does not allow us to extrapolate towards negative outcomes.

This work should benefit the research community of language emergence and grounding, in its current state.

\begin{ack}
This work was supported by the EPSRC Centre for Doctoral Training in Intelligent Games \& Games Intelligence (IGGI) [EP/L015846/1]. 

We gratefully acknowledge the use of Python\cite{python-2009}, IPython\cite{ipython-perez-2007}, SciPy\cite{SciPy-NMeth2020}, Scikit-learn\cite{Scikit-learn:JMLR:v12:pedregosa11a}, Scikit-image\cite{scikit-image-van2014}, NumPy\cite{NumPy-Array2020}, Pandas\cite{pandas1-mckinney-proc-scipy-2010,pandas2-reback2020}, OpenCV\cite{opencv_library}, PyTorch\cite{pytorch-paszke-NEURIPS2019_9015}, TensorboardX\cite{huang2018tensorboardx}, Tensorboard from the Tensorflow ecosystem\cite{tensorflow2015-whitepaper}, without which this work would not be possible.
\end{ack}

\bibliography{bibli}

\section*{Checklist}

The checklist follows the references.  Please
read the checklist guidelines carefully for information on how to answer these
questions.  For each question, change the default \answerTODO{} to \answerYes{},
\answerNo{}, or \answerNA{}.  You are strongly encouraged to include a {\bf
justification to your answer}, either by referencing the appropriate section of
your paper or providing a brief inline description.  For example:
\begin{itemize}
  \item Did you include the license to the code and datasets? \answerYes{See Section~\ref{gen_inst}.}
  \item Did you include the license to the code and datasets? \answerNo{The code and the data are proprietary.}
  \item Did you include the license to the code and datasets? \answerNA{}
\end{itemize}
Please do not modify the questions and only use the provided macros for your
answers.  Note that the Checklist section does not count towards the page
limit.  In your paper, please delete this instructions block and only keep the
Checklist section heading above along with the questions/answers below.

\begin{enumerate}

\item For all authors...
\begin{enumerate}
  \item Do the main claims made in the abstract and introduction accurately reflect the paper's contributions and scope?
    \answerYes{3 main contributions summarised in the abstract and introduction and discussed in details in the result subsections.}
  \item Did you describe the limitations of your work?
    \answerYes{Constrained to visual referential games with specific approaches.}
  \item Did you discuss any potential negative societal impacts of your work?
    \answerYes{In the broader impact section.}
  \item Have you read the ethics review guidelines and ensured that your paper conforms to them?
    \answerYes{}
\end{enumerate}

\item If you are including theoretical results...
\begin{enumerate}
  \item Did you state the full set of assumptions of all theoretical results?
    \answerNA{}
	\item Did you include complete proofs of all theoretical results?
    \answerNA{}
\end{enumerate}

\item If you ran experiments...
\begin{enumerate}
  \item Did you include the code, data, and instructions needed to reproduce the main experimental results (either in the supplemental material or as a URL)?
    \answerYes{Both in the main paper, and as an open-sourced code.}
  \item Did you specify all the training details (e.g., data splits, hyperparameters, how they were chosen)?
    \answerYes{In the experimental setup section.}
	\item Did you report error bars (e.g., with respect to the random seed after running experiments multiple times)?
    \answerYes{All plots contain some measure of uncertainty.}
    \item Did you include the total amount of compute and the type of resources used (e.g., type of GPUs, internal cluster, or cloud provider)?
    \answerNo{It will be added once the review process will have highlighted what is necessary in the paper from what is superfluous, and maybe what other experiments should be ran...}
\end{enumerate}

\item If you are using existing assets (e.g., code, data, models) or curating/releasing new assets...
\begin{enumerate}
  \item If your work uses existing assets, did you cite the creators?
    \answerNA{}
  \item Did you mention the license of the assets?
    \answerNA{}
  \item Did you include any new assets either in the supplemental material or as a URL?
    \answerNA{}
  \item Did you discuss whether and how consent was obtained from people whose data you're using/curating?
    \answerNA{}
  \item Did you discuss whether the data you are using/curating contains personally identifiable information or offensive content?
    \answerNA{}
\end{enumerate}

\item If you used crowdsourcing or conducted research with human subjects...
\begin{enumerate}
  \item Did you include the full text of instructions given to participants and screenshots, if applicable?
    \answerNA{}
  \item Did you describe any potential participant risks, with links to Institutional Review Board (IRB) approvals, if applicable?
    \answerNA{}
  \item Did you include the estimated hourly wage paid to participants and the total amount spent on participant compensation?
    \answerNA{}
\end{enumerate}

\end{enumerate}


\appendix
\section{Appendix A : Spearman $\rho$ Correlation Matrices}
\label{sec:appendixA}

\begin{figure}[h]
    \centering
    \begin{subfigure}[t]{1.0\linewidth}
        \includegraphics[width=1.0\linewidth]{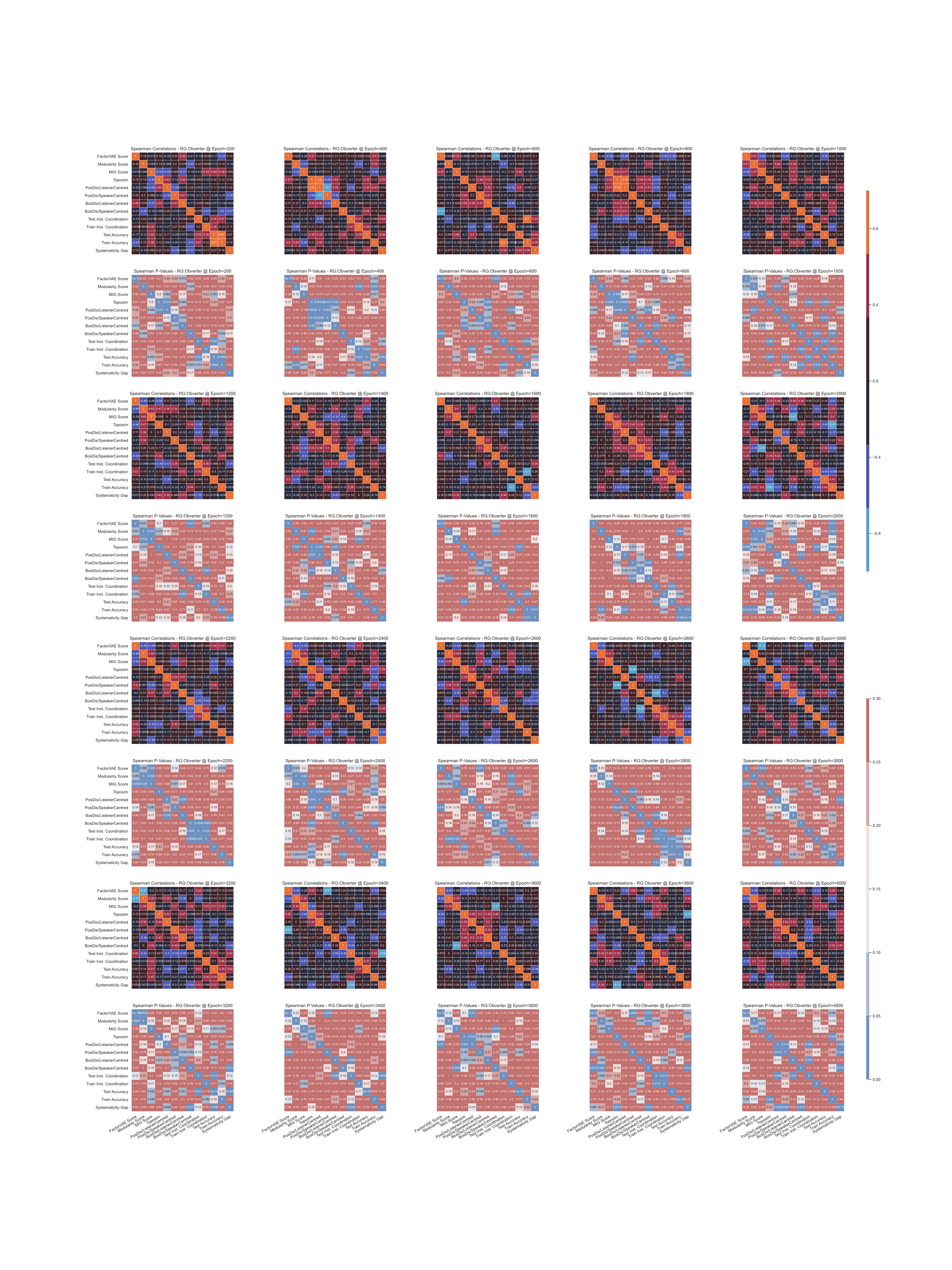}
    \end{subfigure}
    \caption{
    Spearman $\rho$ correlation coefficients (odd rows) and p-values (even rows) matrices between the different metrics, from epoch $200$ to $4000$, with an interval of $200$ epochs between measures, in the context of \textbf{RG:Obverter}. }
    \label{fig:spearman-correlation-matrix-3dshapes-low-RG-Obverter}
\end{figure}

\begin{figure}[h]
    \centering
    \begin{subfigure}[t]{1.0\linewidth}
        \includegraphics[width=1.0\linewidth]{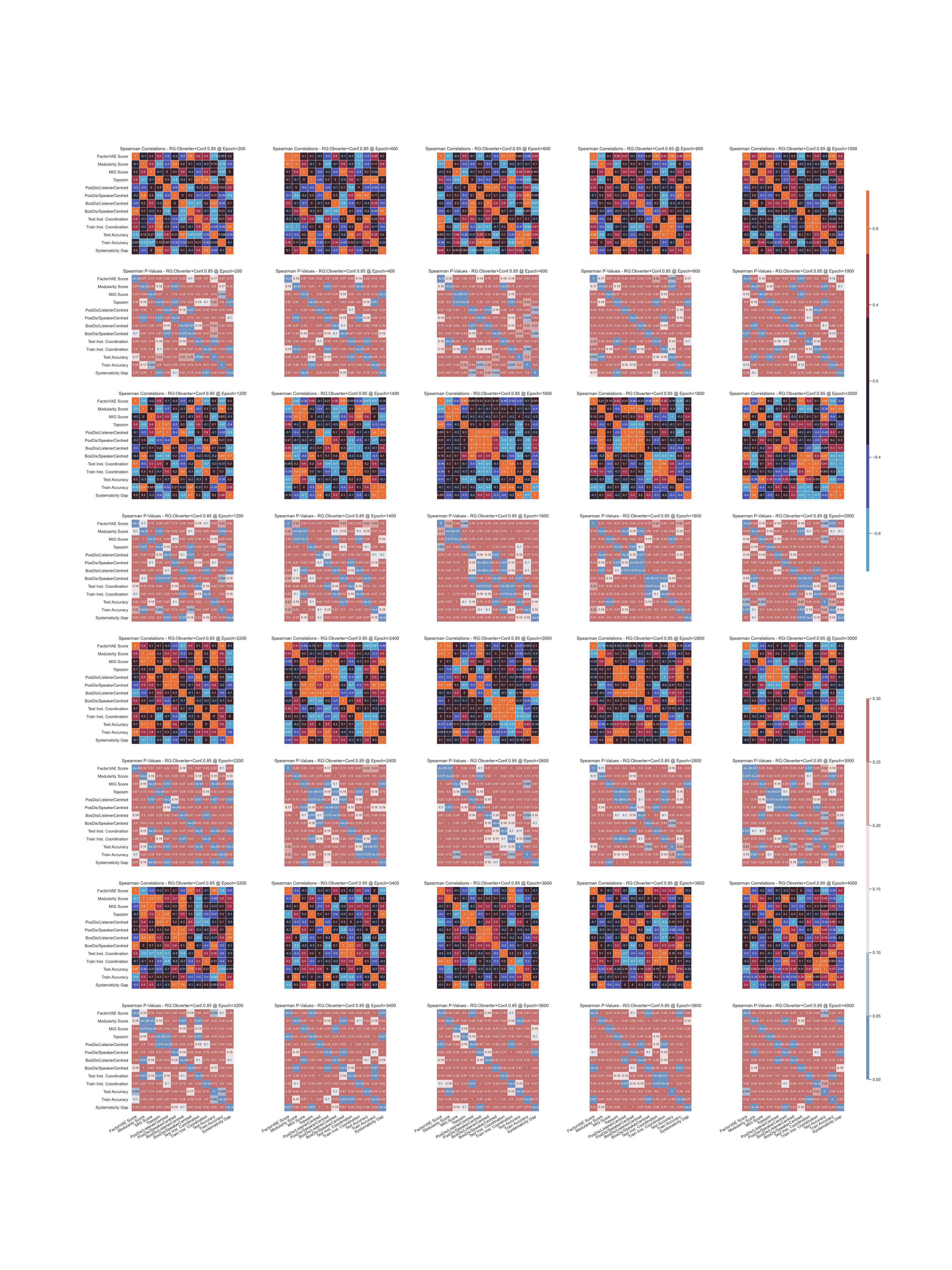}
    \end{subfigure}
    \caption{
    Spearman $\rho$ correlation coefficients (odd rows) and p-values (even rows) matrices between the different metrics, from epoch $200$ to $4000$, with an interval of $200$ epochs between measures, in the context of \textbf{RG:Obverter+Conf.0.85}. }
    \label{fig:spearman-correlation-matrix-3dshapes-low-RG-Obverter-Conf0.85}
\end{figure}

\begin{figure}[h]
    \centering
    \begin{subfigure}[t]{1.0\linewidth}
        \includegraphics[width=1.0\linewidth]{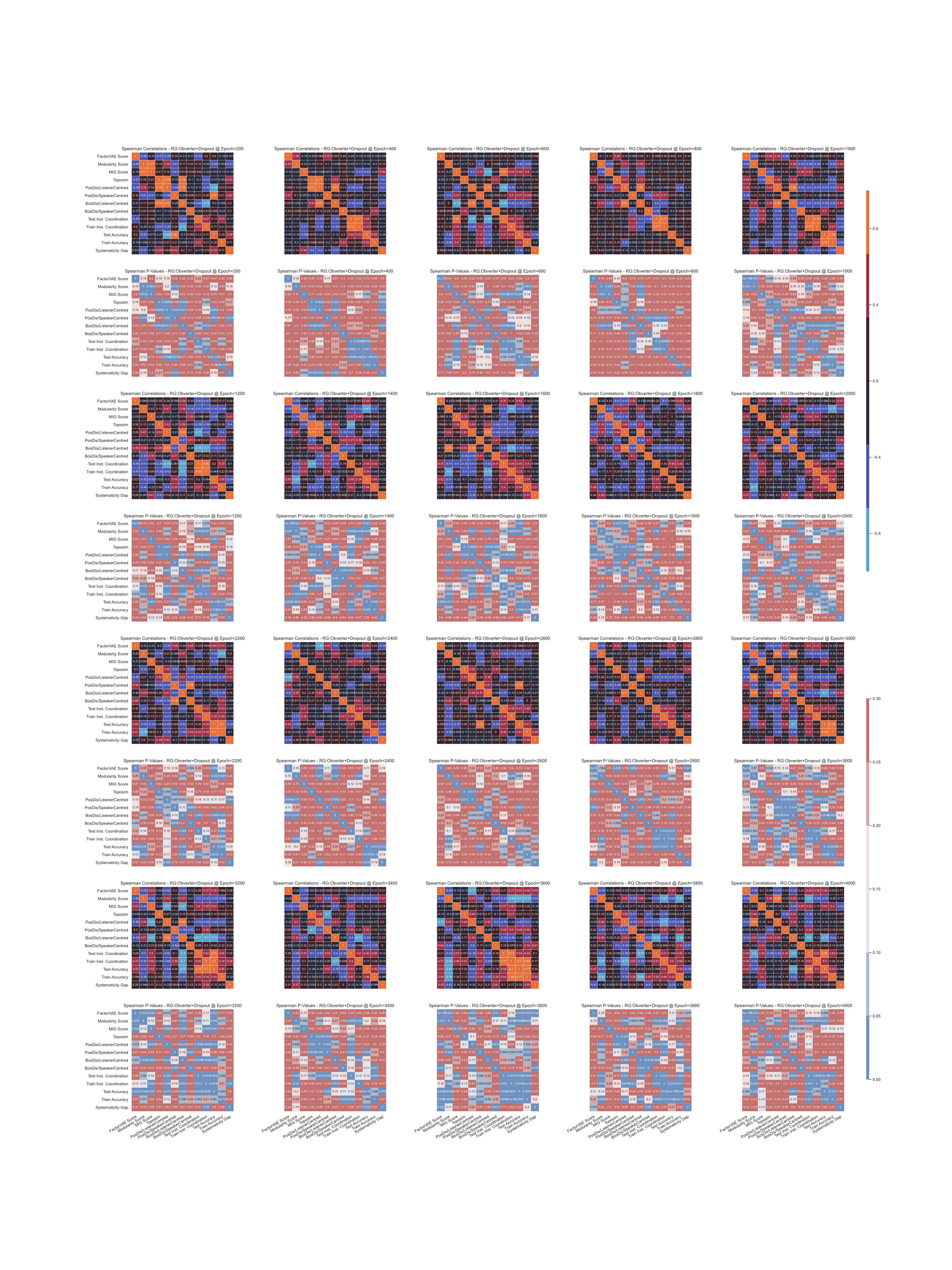}
    \end{subfigure}
    \caption{
    Spearman $\rho$ correlation coefficients (odd rows) and p-values (even rows) matrices between the different metrics, from epoch $200$ to $4000$, with an interval of $200$ epochs between measures, in the context of \textbf{RG:Obverter+Dropout}. }
    \label{fig:spearman-correlation-matrix-3dshapes-low-RG-Obverter-Dropout}
\end{figure}

\end{document}